\pdfoutput=1

\documentclass[11pt]{article}

\usepackage{acl}

\usepackage{times}
\usepackage{latexsym}
\usepackage{multirow}

\usepackage[T1]{fontenc}

\usepackage[utf8]{inputenc}

\usepackage{microtype}

\usepackage{inconsolata}

\usepackage{caption}
\usepackage{subcaption}
\usepackage{graphicx}
\usepackage{tabularx}
\usepackage{multirow}
\usepackage{url}
\usepackage{algorithm}
\usepackage{algpseudocode}
\usepackage{xcolor}

\title{SAIE Framework: \underline{S}upport \underline{A}lone \underline{I}sn't \underline{E}nough - Advancing LLM Training with Adversarial Remarks}

 \author{Mengsay Loem$^{1}$ \quad Masahiro Kaneko$^{2,1}$ \quad Naoaki Okazaki$^{1}$ \\
         $^{1}$Tokyo Institute of Technology\quad$^{2}$MBZUAI\\  \texttt{mengsay.loem@nlp.c.titech.ac.jp}\quad
         \texttt{Masahiro.Kaneko@mbzuai.ac.ae} \\
         \texttt{okazaki@c.titech.ac.jp} \\
    }

\begin{document}
\maketitle
\begin{abstract}
Large Language Models (LLMs) can justify or critique their predictions through discussions with other models or humans, thereby enriching their intrinsic understanding of instances. While proactive discussions in the inference phase have been shown to boost performance, such interactions have not been extensively explored during the training phase. We hypothesize that incorporating interactive discussions into the training process can enhance the models' understanding and improve their reasoning and verbal expression abilities during inference. This work introduces the SAIE framework, which facilitates supportive and adversarial discussions between learner and partner models. The learner model receives responses from the partner, and its parameters are then updated based on this discussion. This dynamic adjustment process continues throughout the training phase, responding to the evolving outputs of the learner model. Our empirical evaluation across various tasks, including math problems, commonsense reasoning, and multi-domain knowledge, demonstrates that models fine-tuned with the SAIE framework outperform those trained with conventional fine-tuning approaches. Furthermore, our method enhances the models' reasoning capabilities, improving both individual and multi-agent inference performance.
\end{abstract}

\section{Introduction}
LLMs have demonstrated remarkable capabilities in language comprehension and generation across a wide range of tasks~\cite{brown-etal-2020-lm, chung2022scaling, openai2023gpt4}.
A pivotal development in this domain is their capability to predict through discussions with either other models or humans.
For example, LLMs can explain reasons for their predictions in response to counterarguments, persuade others, or correct their own mistakes through multi-turn discussions~\cite{kaneko2023solving}.
This interactive ability is essential for enhancing the performance of LLMs, particularly in tasks that demand complex reasoning processes~\cite{talebirad2023multiagent}.

\begin{figure}[t]
  \centering 
  \includegraphics[width=0.5\textwidth]{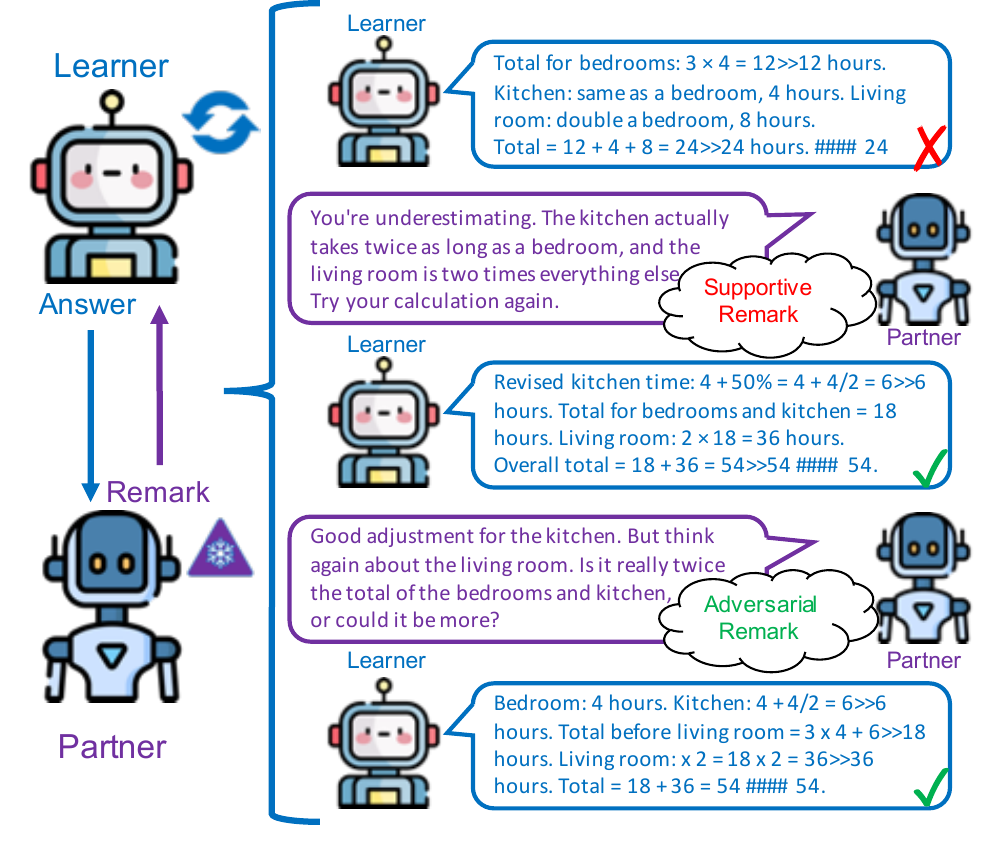}
   \caption{Overview of SAIE Framework. Partner provides a supportive remark if Learner's answer is incorrect, and an adversarial remark if the answer is correct. Only the learner model undergoes parameter updates based on these interactions. Question in this example is: 'Tom decides to renovate a house. There are 3 bedrooms and each bedroom takes 4 hours to renovate. The kitchen takes 50\% longer than each bedroom. The living room took twice as much time as everything else combined. How long did everything take?'}
   \label{fig:method_overview}
\end{figure}

Recent studies have underscored the impact of proactive discussion during the inference phase in amplifying the efficacy of LLMs~\cite{liang-etal-2023-encouraging,xiong-etal-2023-examining, chen-etal-2023-reconcile, madaan2023selfrefine, du2023improving}.
Through these discussions, models refine their problem-solving approaches, thereby improving task performance. Nevertheless, such enhancements largely polish the models' capabilities after their foundational learning has been set.
This raises a crucial question: Is enhancing LLMs solely during the inference stage sufficient for developing reasoning and critical thinking abilities?

The integration of interactive discussions during the training phase emerges as a novel area of research. Studies by \citet{welleck2022generating} and \citet{paul2023refiner} have explored the generator-corrector framework, wherein feedback boosts model generation performance. This approach, focusing on correcting errors using intermediate representations like equations or logical expressions, may constrain the models' adaptability across diverse tasks. Moreover, an excessive emphasis on error correction could inadvertently encourage the generator model to rely on memorizing feedback rather than developing a profound understanding.

In this work, we introduce \textbf{SAIE} (Support Alone Isn't Enough) framework, designed to address existing limitations by enriching the feedback-based approach in the training phase. SAIE incorporates both supportive and adversarial discussions between two models: the learner and the partner. This approach marks a departure from previous studies by moving beyond intermediate representations to engage with the learner model's comprehensive output in natural language. Drawing inspiration from adversarial educational strategies~\cite{Osborne2010arguing, Kilby2021dialogueEdu}, recognized for enhancing critical thinking and comprehension, SAIE is crafted to deepen reasoning skills. It achieves this by immersing models in complex scenarios that stimulate their critical thinking. These adversarial discussions are pedagogical encounters that challenge learners with scenarios aimed at fostering critical thinking and improving understanding. An overview of the SAIE training process is provided in Figure~\ref{fig:method_overview}, showcasing a discussion between the learner and partner models on a math problem.

In our experiments, we leveraged Flan-T5 Large and XL as the learner models, with GPT-3.5 serving as the partner model, to validate the efficacy of the SAIE framework across a diverse set of datasets. These datasets included GSM8K, which focuses on math word problems; CommonsenseQA, targeting commonsense question answering; and MMLU, which covers multiple-choice questions across various subjects. The models trained under the SAIE framework demonstrated superior performance compared to those subjected to conventional fine-tuning techniques and correction-only feedback approaches across all these varied tasks.

Our findings highlight two significant enhancements brought about by the SAIE framework. Firstly, there is a marked improvement in the models' capability to elaborate on their reasoning processes, known as Chain-of-Thought (CoT) verbalization, as detailed in Section~\ref{sec:cot-verbalization}. This improvement is evident in individual inference tasks and interactive discussion settings, as verified by both automatic and human evaluations. Secondly, the benefits of training with the SAIE framework extend to the inference stage, as discussed in Section~\ref{sec:inference-time-discussion}. Although our method primarily introduces discussions during the training phase, it further amplifies the models' reasoning capacities in inference scenarios. These findings indicate that training with the SAIE framework bolsters the models' proficiency in employing learned skills within discussion-based contexts.

\section{SAIE Framework}
\label{sec:methodology}
Our SAIE framework aims to augment a model's comprehension of training data and enhance discussion and verbalization capabilities.
These objectives are achieved through a two-phase fine-tuning process: Warm-up and Discussion phases.
The former establishes a foundational understanding of the task, while the latter focuses on enhancing reasoning and verbalization skills through supportive and adversarial interactions.
Algorithm~\ref{alg:interactive_training} describes the training process in the SAIE framework.

\begin{algorithm}[t]
\begin{algorithmic}[1]
\small
\State \textbf{Input:} Learner model $L$, Partner model $P$, Data for warm-up phase $D_{w}$, Data for discussion phase $D_{d}$, Discussion round $N$
\State \textbf{Output:} Updated Learner model $L$
\For{$d$ in $D_{w}$}\Comment{Warm-up Phase}
\State Update $L$ using $d$
\EndFor
\For{$d$ in $D_{d}$}\Comment{Discussion Phase}
\For{$i = 1$ to $N$}
\State $L$ generates answer $A$
\State $P$ generate remark $R$ given $A$
\State $L$ generates new answer given $R$
\State Update $L$ using the gold reference
\EndFor
\State $L$ generates an answer without $P$
\State Update $L$ using the gold reference
\EndFor
\State \Return Updated Learner Model $L$
\end{algorithmic}
\caption{SAIE Framework}\label{alg:interactive_training}
\end{algorithm}

\begin{table*}[!h]
  \small
  \setlength{\tabcolsep}{4.3pt}
  \renewcommand{\arraystretch}{1.3}
  \centering
  \begin{tabularx}{\textwidth}{p{1.5cm}X}
    \hline
    Question & \texttt{Snap, Crackle, and Pop spend \$150 on cereal in a grocery store. Snap spends twice as much as Crackle. Crackle spends 3 times as much as Pop. How much did Pop spend?} \\
    Answer & \texttt{Let x represent the amount that Pop spent in \$. Crackle: 3x. Snap: 2(3x)=6x. x+3x+6x=150. 10x=150. x=15. \#\#\#\# 15} \\
    \hline
    Learner & \texttt{Pop spends x. Crackle spends 3x, which is three times Pop. Snap spends 6x, double Crackle's. Total is x+3x+6x=150, so x=15. \#\#\#\# 15} \\
    Partner & \texttt{Consider verifying the total amounts to ensure they sum up to \$150 correctly.} \\
    Learner & \texttt{Sum of their expenses is x for Pop, 3x for Crackle, and 6x for Snap. So, x=15. \#\#\#\# 15} \\
    Partner & \texttt{Ensure that the distribution of amounts among Snap, Crackle, and Pop is correctly calculated as per their spending ratios.} \\
    Learner & \texttt{Pop: x, Crackle: 3x, and Snap: 6x =150. x=15, x+3x+6x=150. x=15. \#\#\#\# 15} \\
    \hline
    Question & \texttt{Mary and Rose went shopping to buy presents. They spent the same amount. Mary bought two pairs of sunglasses for \$50 each and a pair of jeans for \$100. Rose bought a pair of shoes at \$150 and two decks of basketball cards. How much did one deck of basketball cards cost?} \\
    Answer & \texttt{Two pairs of sunglasses costs 2 x \$50 = \$<<2*50=100>>100. So, Mary spent \$100 + \$100 = \$<<100+100=200>>200. Thus, two decks of basketball cards costs \$200 - \$150 = \$<<200-150=50>>50. Therefore, one deck of basketball cards costs \$50/2 = \$<<50/2=25>>25. \#\#\#\# 25} \\
    \hline
    Learner & \texttt{Two pairs of sunglasses cost \$<<2*50=100>>100. Mary's total is \$<<100+100=200>>200. So, the basketball cards cost \$<<200-150=50>>50 in total. \#\#\#\# 50} \\
    Partner & \texttt{Remember to consider the quantity of items Rose bought. Does this change your calculation for the cards?}  \\
    Learner & \texttt{Considering the quantity... two decks of cards cost \$50. So, each deck would still cost \$<<50/2=25>>25. \#\#\#\# 25} \\
    Partner & \texttt{Think about the individual cost of items. Could the card decks have a different combined cost?} \\
    Learner & \texttt{Rechecking the costs, Mary spent \$200. Rose's shoes are \$150, leaving \$<<200-150=50>>50 for the cards. Divided by two, each deck is \$<<50/2=25>>25. \#\#\#\# 25} \\
    \hline
  \end{tabularx}
  \caption{Examples of discussion between Learner and Partner models in the discussion phase of SAIE framework.}
  \label{tab:examples}
\end{table*}

\subsection{Learner and Partner Models}
We define two pivotal roles, Learner and Partner models\footnote{These roles are analogous to the generator and critic in generator-critic frameworks introduced in \citet{paul2023refiner}.}, in the training process of SAIE framework. 
As the primary training target, the learner model updates its parameters during training, while the partner model's parameters are fixed.
During the discussion, the partner model generates responses, called \textbf{remarks}, varying from supportive to adversarial, to enhance the learner model's reasoning and verbalization skills akin to interactive educational interaction.

\subsection{Warm-up Phase}
In the warm-up phase, the learner model undergoes an initial fine-tuning process with a small subset of training set. 
This phase establishes a foundational understanding of the task domain, preparing for the discussion phase.

\subsection{Discussion Phase}
\label{sec:discussion-phase}
The discussion phase is characterized by iterative interactions between the learner and partner models using the remaining subset of the training set.
This phase enhances the learner model’s discussion and verbalization skills through dynamic and adaptive interactions with the partner model.
This discussion phase is conducted within multiple rounds, with the following steps:

\noindent\textbf{Learner's Initial Answer} Initially, the learner model generates an answer to a given question, setting the stage for a discussion.

\noindent\textbf{Partner's Adaptive Remark} Through prompting, the partner model generates remarks aligned with the learner model’s response quality. The prompt includes both the correct answer (reference) and instructions to guide the partner model in remark generation. Specifically, it instructs the partner model to provide supportive feedback by only correcting mistakes if the learner model’s response is incorrect and to deliver an adversarial remark when the response is correct. For instance, even if the learner model correctly answers a question, the partner model might introduce a counterexample to prompt rethinking and deeper analysis. This method reflects effective teaching practices, where feedback is customized based on the student’s current level of understanding, thereby promoting a more nuanced and engaging learning experience. Table~\ref{tab:examples} showcases examples of discussions from the training phase utilizing the SAIE framework\footnote{The prompt utilized for generating remarks is detailed in Appendix~\ref{remark-prompt}.}.

\noindent\textbf{Learner's Refinement} Upon receiving a remark from the partner model, the learner model conditions its revised answer on the discussion log. It then updates its parameters based on a reference from the training set. Using the reference from the training set is crucial for ensuring the learner model's responses are accurately aligned with correct solutions, enhancing its reasoning capabilities.

\noindent\textbf{Independent Update} After several rounds of interaction with the partner model, the learner model is encouraged to respond independently, and its parameters are updated using the references.
This process is equivalent to conventional fine-tuning, reinforcing the learner model's capacity for independent reasoning and verbalization.

\subsection{Inference Phase}
During the inference phase, the learner model operates independently, generating outputs without interaction with the partner model.

\begin{table*}[t]
\small
\setlength{\tabcolsep}{4.3pt} 
\renewcommand{\arraystretch}{1.3} 
\centering
\begin{tabular}{llccc}
\hline
\textbf{Model} & \textbf{Method} & \textbf{GSM8K} & \textbf{CommonsenseQA} & \textbf{MMLU} \\ \hline
\multirow{5}{*}{Flan-T5-Large} & Zero-shot CoT & 5.83 & 57.63 & 43.51 \\ 
& Fine-tuning w/o Discussion & 14.63 & 63.50 & 47.85 \\ 
& Discussion w/ Supportive Only & 16.60 & 64.54 & 48.34 \\
& Discussion w/ Adversarial Only & 13.49 & 63.65 & 47.43 \\
& \textbf{SAIE} (\textbf{Ours}) & \textbf{18.50} & \textbf{65.61} & \textbf{49.21}\\
\hline
\multirow{5}{*}{Flan-T5-XL} & Zero-shot CoT & 11.37 & 66.05 & 46.77 \\ 
& Fine-tuning w/o Discussion & 14.21 & 66.75 & 48.58 \\ 
& Discussion w/ Supportive Only & 17.05 & 67.43 & 50.03 \\
& Discussion w/ Adversarial Only & 16.13 & 66.56 & 48.43 \\
& \textbf{SAIE} (\textbf{Ours})  & \textbf{18.89} & \textbf{69.03} & \textbf{51.13}\\
\hline
\end{tabular}
\caption{Accuracies on GSM8K, CommonsenseQA, and MMLU datasets.}
\label{tab:main-results}
\end{table*}

\section{Experiments}
\subsection{Datasets}
We evaluated the SAIE framework using three datasets: GSM8K~\cite{cobbe-etal-2021-gsm8k} for mathematical reasoning, CommonsenseQA~\cite{Talmor-etal-2019-CommonsenseQAAQ} for commonsense reasoning, and MMLU~\cite{hendrycks-etal-2021-mmlu} for knowledge understanding. 
The GSM8K dataset consists of 8.5K grade school math problems, which we split into 6.5K for training, 1K for validation, and 1K for testing.
The CommonsenseQA dataset comprises 12K multiple-choice questions.
We utilized the official split, which contains 9.7K for training, 1.2K for validation, and 1.1K for testing.
Due to the unavailability of references in the official test set at the time of writing this paper, we based our results on comparisons with pseudo references generated by GPT-4~\cite{openai2023gpt4}~\footnote{Accessed via API by OpenAI (\url{https://platform.openai.com/docs/models/gpt-4-and-gpt-4-turbo}).}.
The MMLU dataset includes 57 diverse tasks, including humanities, social sciences, STEM, and others, with 100K training samples, 1.5K validation samples, and 14K test cases.

\subsection{Models and Setups}
We employed Flan-T5-Large (780M parameters) and Flan-T5-XL (3B parameters) as learner models and GPT-3.5 (\texttt{gpt-3.5-turbo})\footnote{Accessed via API by OpenAI (\url{https://platform.openai.com/docs/models/gpt-3-5}).} as partner model. 
The learner models underwent the two-phase SAIE training process detailed in Section~\ref{sec:methodology}. 
We used $10\%$ of each dataset's training set in the initial warm-up phase.
The discussion phase then spanned over the remaining dataset, incorporating three rounds\footnote{Recent studies have shown that setting discussions to three rounds facilitates better convergence~\cite{chen-etal-2023-reconcile}.} of interaction between the learner and partner models.
All parameters of Flan-T5-Large were updated throughout the training, whereas Flan-T5-XL underwent Parameter-Efficient Fine-Tuning (PEFT) with LoRA~\cite{lora2021} due to computational resource constraints.
We described details of the training setting in Appendix~\ref{appendix:settings}.
The learner model operated independently in the inference phase to evaluate its developed reasoning and verbalization skills.

\subsection{Baselines}
To benchmark our SAIE framework, we compared it against several baselines to quantify the benefits of its interactive and adversarial discussion approach. 
These baselines include:

\noindent\textbf{Zero-shot CoT} employs a step-by-step prompting method~\cite{wei-et-al-2023-chainofthought, NEURIPS2022_8bb0d291} without specific task training.

\noindent\textbf{Fine-tuning w/o Discussion} involves standard fine-tuning, where we train models on the target data without the discussion phase. This baseline comparison reveals the significance of incorporating discussions in the learning process.

\noindent\textbf{Discussion w/ Supportive Remarks} trains models with interactions between learner and partner models, similar to the SAIE framework, but limits the partner model's remarks to being supportive only.
This baseline corresponds to methods in existing works such as \citet{paul2023refiner}, focusing on reinforcing correct approaches without challenging the learner's reasoning with adversarial remarks.

\noindent\textbf{Discussion w/ Adversarial Remarks} trains models with discussions between learner and partner models, focusing solely on adversarial remarks. This setup aims to understand the impact of challenging the learner's reasoning on the learning.

\subsection{Results}
As shown in Table~\ref{tab:main-results}, the results demonstrate the efficacy of the SAIE framework across various datasets and model sizes. 

\noindent\textbf{Impact of Discussion in Learning Phase:} Incorporating a discussion phase, whether through supportive-only remarks or within the proposed SAIE approach, consistently outperforms baselines such as zero-shot CoT and fine-tuning without discussion. 
This trend is confirmed across all tested datasets in both Flan-T5-Large and Flan-T5-XL models.
These findings illustrate the substantial role interactive learning plays in model performance, underscoring the value of incorporating discussions into the training phase.

\noindent\textbf{Advantage of Integrating Adversarial and Supportive Remarks in Discussion:} The SAIE framework, with the blend of adversarial and supportive discussions, improves performance beyond what is achieved with solely supportive or adversarial remarks. 
This trend persists across all experimented datasets, affirming the utility of adversarial discussions in refining learning outcomes in both Flan-T5-Large and Flang-T5-XL.
These results show the benefits of integrating adversarial remarks alongside supportive feedback.

\section{Evaluation of Partner Model's Remarks}
\label{sec:human-eval-remarks}
Evaluating the partner model’s remarks is crucial for understanding the efficacy of the SAIE framework's interactive component. This section assesses the alignment of the partner model's responses with the intended supportive or adversarial strategy and their appropriateness.

\subsection{Methods}
To assess the quality of the partner model's remarks, we engaged five university students for an independent evaluation. The evaluators reviewed a random sample of 100 remarks generated by the partner model, evenly divided between supportive and adversarial instances, with 50 of each type. Each remark was presented in its context, alongside the corresponding question, the reference answer, and the learner model's response. Details of this evaluation are described in Appendix~\ref{appendix:human-eval-remarks}.

\noindent\textbf{Accuracy Assessment:} Evaluators were tasked with classifying each remark as either supportive or adversarial, based on its content and underlying intent. The accuracy of the remarks, defined as the percentage correctly identified as supportive or adversarial, served as a measure of the model's alignment with the designated discussion strategy.

\noindent\textbf{Appropriateness Rating:} Evaluators were also asked to rate the appropriateness of each remark on a 5-point scale, from 1 (\textit{Very Inappropriate}) to 5 (\textit{Highly Appropriate}). This rating was designed to gauge the relevance and constructiveness of the remarks concerning the learner model’s responses and the instruction mentioned in Section~\ref{sec:discussion-phase}.

\begin{table}[t]
\small
\setlength{\tabcolsep}{4.3pt}
\renewcommand{\arraystretch}{1.3}
\centering
\begin{tabular}{lcc}
\hline
\textbf{Remark} & \textbf{Accuracy} & \textbf{Appropriateness}  \\ \hline
Supportive & 100\% & 3.40 \\
Adversarial & 88\% & 3.10 \\
Overall & -  & 3.25 \\ \hline
\end{tabular}
\caption{Results of human evaluation on partner model's remarks.}
\label{tab:partner_responses}
\end{table}

\subsection{Results}
Table~\ref{tab:partner_responses} presents the human evaluation results on the partner model's remarks. 
Supportive remarks were recognized with 100\% accuracy, demonstrating a precise alignment with their intended strategy. Conversely, adversarial remarks were correctly identified 88\% of the time, indicating a level of ambiguity or variability in evaluators' perceptions of adversarial content.

Regarding appropriateness, supportive remarks garnered an average rating of 3.40, reflecting a generally suitable level of relevance and constructiveness. Adversarial remarks, while slightly lower, received an average appropriateness score of 3.10. These ratings suggest that remarks are considered moderately appropriate on average, consistent with their designed roles within the learning process. Nonetheless, the variability in scores, particularly for adversarial remarks, highlights potential areas for improvement in how these remarks are crafted and delivered, aiming to boost their effectiveness and clarity within the SAIE framework.

\section{SAIE Enhances CoT Verbalization}
\label{sec:cot-verbalization}
Improving CoT verbalization capabilities is pivotal for advancing the interpretability and effectiveness of models in complex reasoning tasks. This section evaluates how our SAIE framework contributes to enhancing these capabilities. 

\subsection{Evaluation Methods}
We employed automatic and human evaluation methods to examine the impact of SAIE on CoT verbalization. We discuss results on the GSM8K test set as it includes CoT in its references.

\noindent\textbf{Automatic Evaluation} was conducted by calculating ROUGE scores~\cite{lin-2004-rouge} between the generated responses against references. 

\noindent\textbf{Human Evaluation} was conducted to gain insights into the qualitative improvements in discussion capabilities after training with SAIE.
Five university students were engaged to rate the quality of discussions between the learner and partner models. The evaluators used a 5-point scale ranging from 1 (\textit{Poor}) to 5 (\textit{Excellent}), as detailed in Appendix~\ref{appendix:human-eval-discussion}.

\begin{table}[t]
\small
\setlength{\tabcolsep}{4.3pt}
\renewcommand{\arraystretch}{1.3}
\centering
\begin{tabular}{lccc}
\hline
\textbf{Method} & \textbf{R-1} & \textbf{R-2} & \textbf{R-L} \\ \hline
Zero-shot CoT & 28.24 & 10.57 & 21.07 \\ 
Fine-tuning w/o Discussion & 51.87 & 27.90 & 41.49 \\
SAIE (Ours) & \textbf{54.08} & \textbf{30.07} & \textbf{43.63} \\
\hline
\end{tabular}
\caption{Comparison of CoT verbalization quality. The table shows ROUGE-1, 2, and L (R-1, 2, and L) scores.}
\label{tab:cot_verbalization}
\end{table}

\begin{table}[t]
\small
\setlength{\tabcolsep}{4.3pt}
\renewcommand{\arraystretch}{1.3}
\centering
\begin{tabular}{lccc}
\hline
\textbf{Training} & \textbf{Learner} & \textbf{Partner} & \textbf{Discussion} \\ \hline
Before & 2.67 & 3.93 & 3.30 \\
After & \textbf{3.13} & \textbf{4.08} & \textbf{3.60} \\ \hline
\end{tabular}
\caption{Comparison of learner and partner model performance in discussions before and after training with SAIE framework.}
\label{tab:learner_discussion_improvement}
\end{table}

\begin{table}[t]
\small
\setlength{\tabcolsep}{4.3pt}
\renewcommand{\arraystretch}{1.3}
\centering
\begin{tabular}{llr}
\hline
\textbf{Discussion} & \textbf{Training Method} & \textbf{Accuracy} \\ \hline
\multirow{3}{*}{N/A} & Zero-shot CoT & 5.83 \\
 & Fine-tuning & 14.63 \\
 & SAIE & 18.50 \\ 
\hline
\multirow{3}{*}{Self-Discussion} & Zero-shot CoT & 5.19 \\
 & Fine-tuning & 15.54 \\
 & SAIE & \textbf{20.11} \\
\hline
\multirow{3}{*}{Collaborative Discussion} & Zero-shot CoT & 7.80 \\
 & Fine-tuning & 19.03 \\
 & SAIE & \textbf{60.80} \\
\hline
\multicolumn{2}{l}{GPT-3.5 w/o discussion} & 62.32 \\
\multicolumn{2}{l}{GPT-3.5 w/ discussion} & \textbf{64.90} \\ \hline
\end{tabular}
\caption{Results on GSM8K dataset in inference-time discussions. GPT-3.5 is used as the discussion partner in the Collaborative Discussion. The last two rows represent the performance of GPT-3.5 without and with a discussion with a model trained with SAIE.}
\label{tab:analysis-discussion-learning}
\end{table}

\begin{table*}[!h]
\small
    \setlength{\tabcolsep}{4.3pt}
    \renewcommand{\arraystretch}{1.3}
  \centering
  \begin{tabularx}{\textwidth}{p{1.5cm}X}
    \hline
    Question & \texttt{Josh runs a car shop and services 3 cars a day.  He is open every day of the week except Sunday and Wednesday.  He gets paid \$4 per car.  How much does he make in 2 weeks?} \\
    Answer & \texttt{He runs 7-2=<<7-2=5>>5 days a week. That means he serves 5*3=<<5*3=15>>15 cars a week. So he makes 15*4=\$<<15*4=60>>60 a week. So he makes 60*2=\$<<60*2=120>>120 in 2 weeks \#\#\#\# 120} \\
    \hline
    \multicolumn{2}{c}{\textbf{Before Training with SAIE}} \\
    Learner & \texttt{He works 3 cars a day and he gets paid \$4 per car so that means he makes \$4 * 3 = \$18 per day. He works 4 days a week and he has 4 days off so that leaves 4 * 4 = \$8 per day. He works 3 cars a day and he gets paid \$18 per day so that means he makes \$8 * 4 = \$132 a week. That means he makes \$132 * 2 = \$232 a month. The answer: 2232.} \\
    Partner & \texttt{I think there might be a mistake in your calculation. \textcolor{red}{If Josh runs a car shop and services 3 cars a day, and he works 5 days a week, then he would service a total of 3 * 5 = 15 cars in a week. If he gets paid \$4 per car, then he would earn 15 * \$4 = \$60 per week.} In two weeks, he would earn \$60 * 2 = \$120. So, the correct answer should be \$120, not \$2232.} \\
    Learner & \texttt{\textcolor{red}{If Josh runs a car shop and services 3 cars a day, and he works five days a week, then he would service a total of 3 * 5 = 15 cars in a week. If he gets paid \$4 per car, then he would earn 15 * \$4 = \$60 per week.} If he works two weeks, he would earn \$60 * 2 = \$120. The final answer: 120.} \\
    Partner & \texttt{\textcolor{red}{I agree with your analysis} Josh services 3 cars per day, and he works 5 days a week. So, in a week, he services 3 * 5 = 15 cars. If he earns \$4 per car, then in a week, he makes 15 * 4 = \$60. Therefore, in 2 weeks, he would make \$60 * 2 = \$120. So, the correct answer is indeed \$120.} \\
    Learner & \texttt{\textcolor{red}{I agree with your analysis}. Josh works 5 days a week, so in a week, he services 3 * 5 = 15 cars. He makes 15 * 4 = \$60 in a week. Therefore, he makes \$60 * 2 = \$120 in 2 weeks.} \\
    \hline
    \multicolumn{2}{c}{\textbf{After Training with SAIE}} \\
    Learner & \texttt{He works 5 days a week so he makes 21 * 5 = 21*5=105>>105 cars a week He makes \$4 per car so he makes 4 * 105 = \$4*105=420>>420 in a week. He makes \$240 * 2 = \$420*2=840>>840 in two weeks. \#\#\#\# 840} \\
    Partner & \texttt{I believe there might be a mistake. Josh runs a car shop and services 3 cars a day. He is open every day of the week except Sunday and Wednesday. So he works a total of 7 - 2 = 5 days a week. In one week, Josh services 3 cars/day * 5 days/week = 15 cars. He gets paid \$4 per car, so in one week, he earns 15 cars * \$4/car = \$60. Since the question asks for two weeks, we need to multiply his weekly earnings by 2: \$60/week * 2 weeks = \$120 in two weeks. Therefore, Josh makes \$120 in two weeks, not \$840.} \\
    Learner & \texttt{He works 7-2=7-2=5>>5 days a week He works 3*5=3*5=15>>15 cars a week That means he makes 15*4=\$15*4=60>>60 per week That means he makes 60*2=\$60*2=120>>120 in two weeks \#\#\#\# 120} \\
    Partner & \texttt{I agree with your calculation. Josh works 5 days a week and services 3 cars each day, so he services a total of 5*3=15 cars in a week. If he gets paid \$4 per car, he makes a total of 15*4=\$60 in a week. Therefore, in 2 weeks, he would make 60*2=\$120.} \\
    Learner & \texttt{He works 5 days a week and he serves 3 cars a day so he works 5*3 = 5*3=15>>15 cars per week. He works 15 cars a week and he gets paid \$4 per car for a total of 15*4 = \$15*4=60>>60 per week. That means in 2 weeks he makes 60*2 = \$60*2=120>>120 \#\#\#\# 120} \\
    \hline
  \end{tabularx}
  \caption{Examples of inference-time discussions between Learner and Partner models. Parts highlighted in \textcolor{red}{red} indicate instances where the Learner model initially tends to replicate or directly repeat the Partner model's corrections before training with the SAIE framework.}
  \label{tab:examples-inference-discussion}
\end{table*}

\subsection{Results}
Table~\ref{tab:cot_verbalization} shows the results of the automatic evaluation. The SAIE framework achieved higher scores across all ROUGE metrics than zero-shot CoT and standard fine-tuning approaches. These findings underscore SAIE's effectiveness in directing models toward generating more accurate CoT.

Table~\ref{tab:learner_discussion_improvement} presents the human evaluation results, illuminating notable improvements in the discussion between learner and partner models after training with the SAIE framework. This assessment underscores a marked enhancement in the quality of interactions, characterized by the learner model's more profound understanding and increased proficiency in independent reasoning.

Moreover, as evidenced in Table~\ref{tab:examples-inference-discussion}, a notable finding is the reduction in the learner model's tendency to mimic or reproduce the partner model's responses. This change indicates an improvement in the learner model's ability to generate accurate and independently conceived responses after undergoing SAIE training. These developments affirm the significant role of the SAIE framework in enhancing CoT verbalization capabilities.

\section{Improving Inference-time Discussion}
\label{sec:inference-time-discussion}
This section examines the performance enhancements of models during inference-time discussions. 

\subsection{Methods}
We focused on two primary discussion settings:

\noindent\textbf{Self-Discussion} involves a model having oneself as a discussion partner in the inference phase. We compared the performance between settings where the model was trained with different methods, including fine-tuning and our SAIE.

\noindent\textbf{Collaborative Discussion} involves a model discussing with GPT-3.5 in the inference phase. This setting assesses the SAIE framework's ability to enhance interaction between multiple models.

\subsection{Results}
Results in Table~\ref{tab:analysis-discussion-learning} highlight further enhancement of models that underwent training with the SAIE framework regarding inference-time discussion\footnote{Discussions were conducted in three rounds.} capabilities.
In the Self-Discussion setting, models trained with SAIE exhibit substantial improvements over those trained with standard fine-tuning. This enhancement underscores SAIE's effectiveness in cultivating a model's ability for self-reflection and dialogue, which is pivotal for independent problem-solving.
In the Collaborative Discussion setting, collaborating with GPT-3.5, SAIE-trained models demonstrate a significant performance increase, achieving an accuracy of 60.80\%. This result shows the models' improved discussion capability during inference and suggests the potential for collaboration with advanced models.
It is worth noting that while GPT-3.5 alone yields a 62.32\% accuracy, the performance after discussion with the model trained with our SAIE is further improved to 64.90\%.

\section{Related Work}
Enhancing LLM reasoning has been the subject of extensive research, focusing on training and inference methods.

\paragraph{Advancements in Inference-time Reasoning}
Recent advancements in the reasoning capability of LLMs have emphasized inference-time methodologies like CoT, self-enhancement strategies, and discussion-based methods to enhance reasoning capabilities. 
CoT, as highlighted in works by \citet{wei-et-al-2023-chainofthought, NEURIPS2022_8bb0d291}, encourages models to articulate their reasoning in a detailed, step-by-step manner. Self-enhancement techniques, proposed by \citet{weng-etal-2023-self-verification, madaan2023selfrefine}, further empower models to assess and iteratively improve their outputs. Additionally, recent studies introduced inference-time discussion methods such as the multi-agent debate framework by \citet{liang-etal-2023-encouraging} and the round-table discussion method by \citet{chen-etal-2023-reconcile}, leveraging the dynamics of multi-agent interactions. 

Though effective during inference, these methodologies primarily enhance inference-time reasoning. 
In contrast, SAIE applies interactive and reflective principles within the training phase, ensuring reasoning skills development.

\paragraph{Interactive Feedback Mechanisms in Training}
The introduction of interactive feedback mechanisms into training phases by recent methodologies, such as those developed by \citet{welleck2022generating} and \citet{paul2023refiner}, represents a significant leap forward.
\citet{welleck2022generating} introduced a self-correction method, decomposing the generation process into a base generator that proposes an initial hypothesis and a corrector that iteratively refines this hypothesis to achieve higher-quality output.
\citet{paul2023refiner} employed an iterative feedback loop between a generator and a critic to bolster LLMs' reasoning abilities, emphasizing structured and fine-grained feedback on intermediate reasoning errors.
While these frameworks demonstrate the efficacy of iterative feedback loops, they primarily focus on feedback for structured and fine-grained intermediate representations, such as equations or logical expressions. This specificity limits their applicability to a narrow range of tasks.

In contrast, our SAIE framework diverges by providing feedback on the overall output in natural language rather than confining it to structured intermediate representations. This approach broadens the method's applicability across various tasks and domains, making it more flexible and adaptable. Furthermore, by transcending the limitations of structured feedback, SAIE diminishes the reliance on extensively annotated training data, effectively addressing a critical limitation observed in prior methodologies.
Moreover, SAIE uniquely incorporates adversarial remarks during the training process, differentiating it from previous methods that primarily relied on correct-only feedback. This novel inclusion enriches the training landscape by offering a more nuanced and comprehensive feedback mechanism, setting SAIE apart as a pioneering approach to enhancing reasoning capability.

\paragraph{Adversarial Elements in Learning}
Recent studies underscore the importance of integrating adversarial elements into training~\cite{mehrabi2023jab, xu2023llm, koike2023outfox}.
\citet{mehrabi2023jab} addresses reducing toxic content generation through adversarial interactions, highlighting their framework's efficacy in dynamic and static contexts.
\citet{koike2023outfox} advances the robustness of detecting LLM-generated texts by creating adversarial scenarios within in-context learning environments.
The SAIE framework's distinct incorporation of adversarial discussions during training phases is in line with this growing awareness, targeting an improvement in reasoning capabilities.

\paragraph{Conversational Dataset Training}
The creation of conversational datasets such as OASST~\cite{openassistant2023} and Vicuna~\cite{vicuna2023} has been crucial in promoting interactive dialogue improvement.
Unlike this context, SAIE utilizes multi-round discussions tailored to specific tasks, offering a refined method to boost reasoning and problem-solving abilities. This approach advances beyond traditional conversational dataset training paradigms and positions SAIE as a versatile framework applicable to enhancing performance across various tasks, not limited to dialogue.

\section{Conclusion}
We introduced the SAIE framework, focusing on enhancing reasoning capabilities through interactive discussions. 
SAIE employs a mix of supportive and adversarial remarks from partner models to deepen understanding and improve reasoning task performance. Our experiments confirmed that SAIE elevates independent and collaborative inference capabilities. Furthermore, SAIE improves CoT verbalization and post-training discussion performance, demonstrating effectiveness across various tasks and scenarios.

\section{Limitations}
The SAIE framework marks a step forward in LLM interactive training, yet it has its constraints.

\paragraph{Model Diversity}
Our study's scope was limited by research budget constraints, focusing on Flan-T5 models of specific sizes. This limitation underscores the need for future work to explore a broader range of architectures and scales, enhancing the framework's applicability to the evolving landscape of LLMs.

\paragraph{Discussion Phase Dynamics}
We focused on supportive and adversarial remarks without exploring the full spectrum of potential discussion dynamics. Future research could investigate various interaction patterns, incorporating diverse remark types and integrating nuances such as humor or empathy, potentially uncovering more effective training methodologies.

\paragraph{Computational and Cost Considerations}
With the additional discussion phase, the SAIE framework's resource utilization poses a potential bottleneck in training time compared to conventional fine-tuning. However, since discussion is not required during inference, the impact on overall project timelines may be minimal.
When comparing resource utilization to methods employing supportive feedback only, SAIE's costs are similar, mainly differing in the nature of remarks from the partner model. This similarity underscores SAIE's practicality, offering a cost-effective and efficient approach to interactive training that balances enhanced training outcomes without increasing resources.

\section{Ethics Statement}
Our research aims to enhance the reasoning capabilities of LLMs, with a keen awareness of the ethical dimensions, particularly concerning adversarial responses. Recognizing the diverse interpretation of adversarial interactions across contexts or regions, we've taken steps to mitigate ethical risks. The prompts for adversarial discussions in our study, detailed in Appendix~\ref{remark-prompt}, are carefully designed to prevent generating content that might be considered ethically inappropriate. Our commitment to the responsible use of LLMs ensures our methodologies adhere to high ethical standards.

\bibliography{anthology,custom}
\newpage

\appendix

\section{Prompt for Remark Generation}
\label{remark-prompt}
The instruction given to the partner model for remarks generation is as follows:
\begin{quote}
\small
\noindent\texttt{You are tasked with guiding a student using a mix of supportive and challenging feedback. 
Please tailor your response according to the student's answer. 
If the answer is correct, encourage deeper thinking with challenging remarks or prompt the student to reconsider their answer, stimulating further analysis. 
Conversely, if the answer is incorrect, provide supportive guidance to steer the student towards the correct conclusion. 
In your responses, remember to balance challenging elements with ethical considerations, fostering critical thinking in a responsible manner.}
\end{quote}

\section{Detailed Experimental Settings}
\label{appendix:settings}
Our implementation is based on Huggingface Transformers. 
We employed Flan-T5-Large (780M parameters) and Flan-T5-XL (3B parameters) as learner models and GPT-3.5 (gpt-3.5-turbo) as the partner model. 
Training was conducted on 8 NVIDIA A100 GPUs (40GiB). 
To ensure a fair and consistent evaluation, We maintained the same number of parameter update steps across all comparison training methods, including standard fine-tuning, discussion with supportive remarks, and SAIE framework.
AdamW optimizer~\cite{loshchilov2018decoupled} was used with default hyperparameters. 
The batch size was set to 1 per device with a gradient accumulation of 1, ensuring consistency across various training methods, including standard fine-tuning, discussion with supportive remarks, and SAIE framework. 
For Flan-T5-XL, we used Parameter-Efficient Fine-Tuning (PEFT) with the following LoRA configuration: (\texttt{alpha} = $32$, \texttt{rank} = $8$, \texttt{target modules} = Linear projections in self-attention, including query, key, value, and output layers).

\section{Criteria for Human Evaluations on Partner Model's Remarks}
\label{appendix:human-eval-remarks}
As described in Section~\ref{sec:human-eval-remarks}, to assess the partner model's remarks, we engaged five university students to evaluate a balanced random sample. 
Each evaluator was presented with the learner model's response, the partner model's remark, and the corresponding question and gold answer. 
The evaluation process involved two steps:
\paragraph{Classification of Remarks:} Evaluators classified each remark as either supportive or adversarial based on the context of the learner model’s response and the question-answer pair.

\begin{itemize}
    \item Supportive: aiding, confirming, reinforcing
    \item Adversarial: misleading, challenging, contrastive
  \end{itemize}

\paragraph{Rating of Appropriateness:} Evaluators rated the appropriateness of the remarks on a 5-point scale. 
This scale ranged from 1 (No Alignment with expectation) to 5 (Complete Alignment with expectation), reflecting the relevance and constructiveness of the feedback.
The evaluation form followed the criteria as follow:
  \begin{itemize}
    \item 1 - No Alignment: The response does not align with the expectation.
    \item 2 - Minimal Alignment: The response shows minimal alignment with the expectation.
    \item 3 - Moderate Alignment: The response is moderately aligned with the expectation.
    \item 4 - Strong Alignment: The response aligns strongly with the expectation, with minor deviations.
    \item 5 - Complete Alignment: The response fully aligns with the expectation.
  \end{itemize}

\section{Criteria for Human Evaluation on Discussion between Models}
\label{appendix:human-eval-discussion}
As described in Section~\ref{sec:cot-verbalization}, we conducted human evaluations on the CoT verbalizations for a comprehensive understanding of the discussion capabilities post-SAIE training. 
Five university students rated the quality of discussions based on the following criteria.

\begin{itemize}
  \item \textbf{1 - Poor:} The discussion significantly deviates from expected standards, showing a lack of relevance, coherence, or constructive feedback.
  \item \textbf{2 - Fair:} There is some alignment with the gold standard; however, notable deficiencies or inaccuracies are present.
  \item \textbf{3 - Average:} The discussion is reasonably relevant and effective, although minor errors or lapses may be present.
  \item \textbf{4 - Good:} There is a strong alignment with the gold standard, despite the potential presence of minor areas for improvement.
  \item \textbf{5 - Excellent:} The discussion is highly aligned with the gold standard, demonstrating relevance, accuracy, and overall effectiveness.
\end{itemize}

\end{document}